\documentclass{article}
\usepackage{amsmath}

\usepackage[final]{neurips_2024}

\usepackage[utf8]{inputenc} 
\usepackage[T1]{fontenc}    
\usepackage{hyperref}       
\usepackage{url}            
\usepackage{booktabs}       
\usepackage{amsfonts}       
\usepackage{nicefrac}       
\usepackage{microtype}      
\usepackage{xcolor}         
\usepackage{graphicx} 
\usepackage{listings}
\usepackage{subcaption}

\definecolor{ForestGreen}{RGB}{34,139,34}

\title{VURF: A General-purpose Reasoning and Self-refinement Framework for Video Understanding}

\author{%
    Ahmad Mahmood$^1$ \\
    amahmood@ethz.ch
    \And 
    Ashmal Vayani$^2$
    \And
    Muzammal Naseer$^3$
    \And
    Salman Khan$^{4,5}$
    \And
    Fahad Shahbaz Khan$^{4,6}$ \vspace{7pt} \\
    $^1$ $\text{ETH Zurich}$ \hspace{10pt}
    $^2$ $\text{University of Central Florida}$ \hspace{10pt}
    $^3$ $\text{Khalifa University, UAE}$ \\
    $^4$ $\text{Mohamed Bin Zayed University of AI, UAE}$ \hspace{10pt}
    $^5$ $\text{Australian National University, Australia}$ \\
    $^6$ $\text{Linköping University, Sweden}$
}

\begin{document}
\maketitle
\begin{abstract}
Recent studies have demonstrated the effectiveness of Large Language Models (LLMs) as reasoning modules that can deconstruct complex tasks into more manageable sub-tasks, particularly when applied to visual reasoning tasks for \emph{images}. In contrast, this paper introduces a \emph{Video Understanding and Reasoning Framework} (VURF) based on the reasoning power of LLMs.  Ours is a novel approach to extend the utility of LLMs in the context of video tasks, leveraging their capacity to generalize from minimal input and output demonstrations within a contextual framework. We harness their contextual learning capabilities by presenting LLMs with pairs of instructions and their corresponding high-level programs to generate executable visual programs for video understanding. 
To enhance the program's accuracy and robustness, we implement two important strategies. \emph{Firstly,} we employ a feedback-generation approach, powered by GPT-3.5, to rectify errors in programs utilizing unsupported functions. \emph{Secondly}, taking motivation from recent works on self-refinement of LLM outputs, we introduce an iterative procedure for improving the quality of the in-context examples by aligning the initial outputs to the outputs that would have been generated had the LLM not been bound by the structure of the in-context examples.  Our results on several video-specific tasks, including visual QA, video anticipation, pose estimation, and multi-video QA, illustrate these enhancements' efficacy in improving the performance of visual programming approaches for video tasks.
\end{abstract}
\section{Introduction}

\label{sec:intro}
In recent years, the vision community has developed highly efficient specialized models for various video understanding tasks, including Video Question Answering \cite{antol2015vqa}, Action Anticipation \cite{girdhar2021anticipative} and Pose Estimation \cite{toshev2014deeppose, koprinska2001temporal, yilmaz2006object, gammulle2019predicting}. Despite such advancements, video models usually offer isolated visual comprehension capabilities in the form of narrow task-specific models. 
Such specialized models limited to individual tasks struggle to offer a comprehensive, adaptable, and scalable understanding of videos for complex reasoning. 
Although the current off-the-shelf models can perform specific well-defined tasks, they require specialized and comprehensive datasets to effectively train dedicated models for each task, which is not feasible for general-purpose complex reasoning problems. 
Furthermore, individual off-the-shelf vision models for each video-understanding task necessitate a distinct framework with unique model configurations. 
This problem underscores the requirement for a uniform reasoning framework that offers plug-and-play architecture, capable of leveraging any pre-trained computer vision model, enabling seamless execution of a given task. 

To address these challenges, we adopt a strategy of decomposing broad video-understanding tasks into more manageable sub-tasks, each of which can be solved by executing task-specific models, and subsequently consolidating the results. 
Our framework is motivated by the observation that complex tasks can be effectively solved by executing an intermediate sequence of sub-tasks, collectively working in sequence to solve the more challenging problem. 
This process of task decomposition necessitates a reasoning module capable of discriminating the necessary steps for task execution. Large Language Models (LLMs) emerge as promising candidates for this role. Recent works on Visual Programming \cite{gupta2023visual} have demonstrated the effectiveness of LLMs in breaking down complex tasks into smaller, more manageable components that can be tackled by specialized computer vision models \cite{wu2023llms,yang2023llm,ruan2023tptu}. In our study, we demonstrate the utility of such a reasoning module in tackling specific challenges within the domain of video understanding. We also show that building such an approach on top of the existing off-the-shelf video models can significantly enhance task performance.

While LLMs demonstrate competence in serving as reasoning modules, they are not immune to errors and limitations. One notable deficiency is their vulnerability to hallucinations induced by contextual information, without the means to self-correct based on task-agnostic knowledge \cite{sun2023head}. To address this concern, we draw inspiration from recent research demonstrating the efficacy of self-refinement processes in enabling LLMs to enhance their outputs \cite{madaan2024self, feng2024improving}, akin to the way humans engage in self-correction. Specifically, we propose a feedback-generation mechanism that evaluates the LLM's output and employs this feedback to prompt the LLM to refine its output. Our findings substantiate the effectiveness of this approach in elevating the performance of video-based reasoning approaches.
Our main contributions are as follows:
\begin{itemize}
    \item The first generic visual reasoning framework for video understanding consolidates multiple task-specific, domain-specialized video models to answer any video-related user queries. 
    \item Using in-context learning, we align LLM behavior for decomposing a given complex task into multiple sub-tasks easily solvable using existing task-specific video models. 
    \item Our proposed self-refinement strategy helps avoid errors in the programs (outlining subtask decomposition) generated by the LLM and boosts the performance by iterative refining the generated program. The proposed framework is shown to boost performance for tasks such as visual question answering for videos in complex reasoning scenarios. 
\end{itemize}

\begin{figure}[t!]  
  \includegraphics[width=\textwidth]{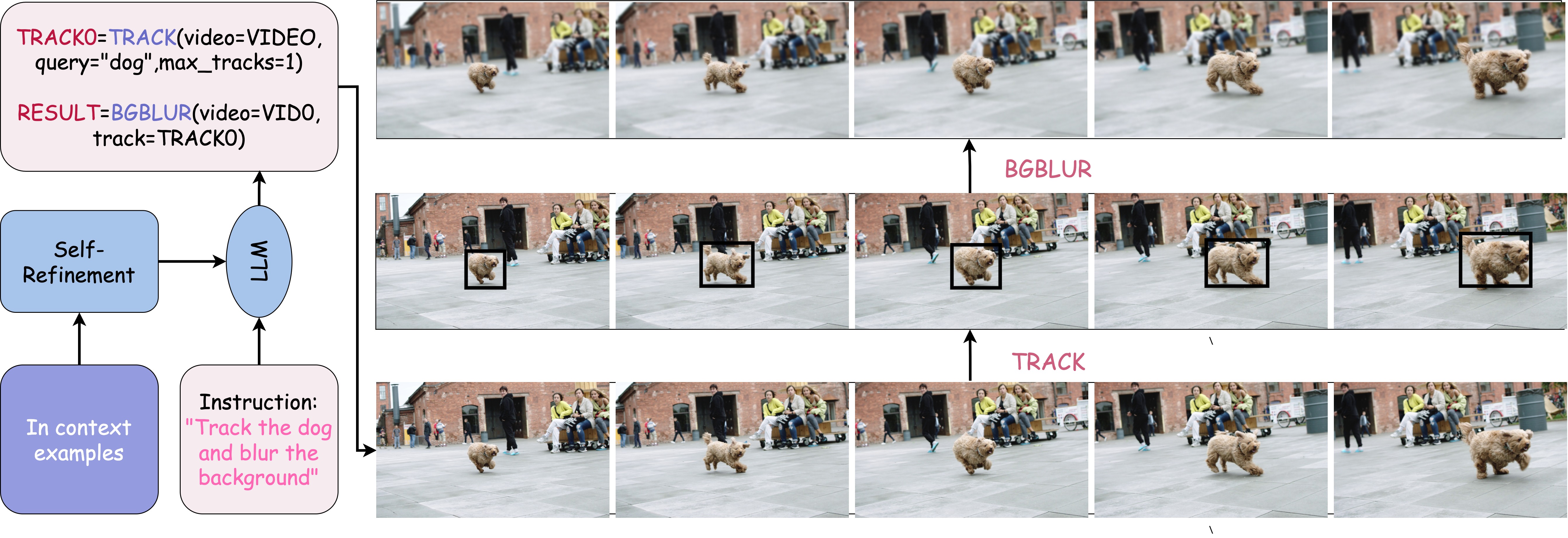}
  \caption{\textbf{An overview of the VURF pipeline:} Figure demonstrates how a complex query regarding video editing is broken down in VURF to arrive at the final edited result.  \textit{Best viewed in zoom.}}
  \label{fig:teaser}
\end{figure}

\section{Related Work}

\textbf{Video Understanding:}
Video understanding focuses on teaching machines to understand and analyze visual content. One crucial task is to recognize and localize different actions in the video \cite{reddy2023synthetic}. Numerous algorithms have been developed to cater the video understanding tasks such as 
Video Swin Transformer \cite{liu2022video}, VideoMAE \cite{tong2022videomae}, C2D \cite{wang2018non}, MViT-V2 \cite{li2022mvitv2}, STGCN++ \cite{yu2017spatio}, and ViViT \cite{arnab2021vivit}
that achieve high accuracy on SOTA datasets \cite{sigurdsson2016hollywood, gao2017tall, gu2018ava, deliege2021soccernet, huang2020movienet, girdhar2021anticipative, sadhu2021visual}.

Video understanding tasks facilitate the efficient handling of diverse information modalities \cite{li2020unimo} and various tasks have been introduced to test the capabilities of the methods for video understanding such as retrieving temporal and spatial information \cite{zhang2023temporal} and answering natural language questions from the video i.e., Video Question Answering (VQA) \cite{yang2003videoqa, lei2018tvqa}. Other methods such as SeViLA \cite{yu2024self} and iVQA \cite{lin2023towards}, adapt Localizer and Answerer for both QA and temporal key-frame localization which are then extended to zero-shot VQA \cite{song2023moviechat, yang2022tubedetr, lin2023univtg}.
Another challenging task in video understanding is to localize the starting and ending time of the video segment that corresponds to the input query i.e., video grounding \cite{chen2018temporally, yang2022tubedetr, zeng2020dense}. It can solve various video understanding tasks such as Temporal Action Recognition \cite{chen2019relation}, spatio-temporal video grounding \cite{zhang2020does}, and Action Recognition \cite{carreira2017quo}. Our work on video programming provides a pipeline for various video understanding and reasoning tasks leveraging off-the-shelf SOTA models and the reasoning capabilities of LLMs for each sub-task.

\textbf{Visual Programming:}
Visual Programming leverages Language Models (LLMs) to break down complex vision-understanding tasks into simpler sub-tasks executed sequentially, improving responses with in-context examples and prompts. Recent advancements, such as VisProg \cite{gupta2023visual}, enhance visual task performance by increasing in-context examples, replacing high-error modules with off-the-shelf models, and refining instructions. Our work, VURF, is the first generic reasoning framework for video, utilizing LLMs' self-critique and refining visual programs to address errors. While zero-shot models like Flamingo \cite{alayrac2022flamingo} can adapt to new tasks without fine-tuning, they struggle with generalization, unlike VURF, which leverages SOTA models for downstream tasks.

\textbf{Self Refinement:}
LLMs demonstrate a special ability to enhance their outputs just like how humans re-evaluate, refine, and reiterate the text that they have initially written. The same LLM when used to identify the potential issues with the output and through the generator, refiner, and feedback provider, it even has the potential to improve responses generated by SOTA LLMs like GPT4 \cite{madaan2023self}. Other methods include using external tools like search engines to rectify the output \cite{madaan2023self}, detecting hallucinated outputs \cite{gou2023critic, evans2021truthful, zhou2020detecting, golovneva2022roscoe}, using natural language feedback \cite{saunders2022self} to improve the initial generated response.

Leveraging the capabilities of self-debugging, language models have been able to debug their predicted program via few-shot demonstrations \cite{chen2023teaching}, use of relevant in-context examples to generate efficient response \cite{wang2023learning, thawakar2024mobillama}, and have contributed in various domains such as code generation and its applications \cite{yu2019cosql}, summarization \cite{campos2022training}, and program synthesis \cite{le2022coderl, kim2023language}.

While these approaches improve the LLM's response via recursive feedback methods, these methods require continuous refinement of any output generated. Our self-refinement approach focuses on the critique and refinement of in-context examples and we show that just a pre-defined set of in-context examples can boost the performance of a visual programming approach (Table \ref{tab:Accuracies}).

\begin{figure}[t!]  
  \includegraphics[width=\textwidth]{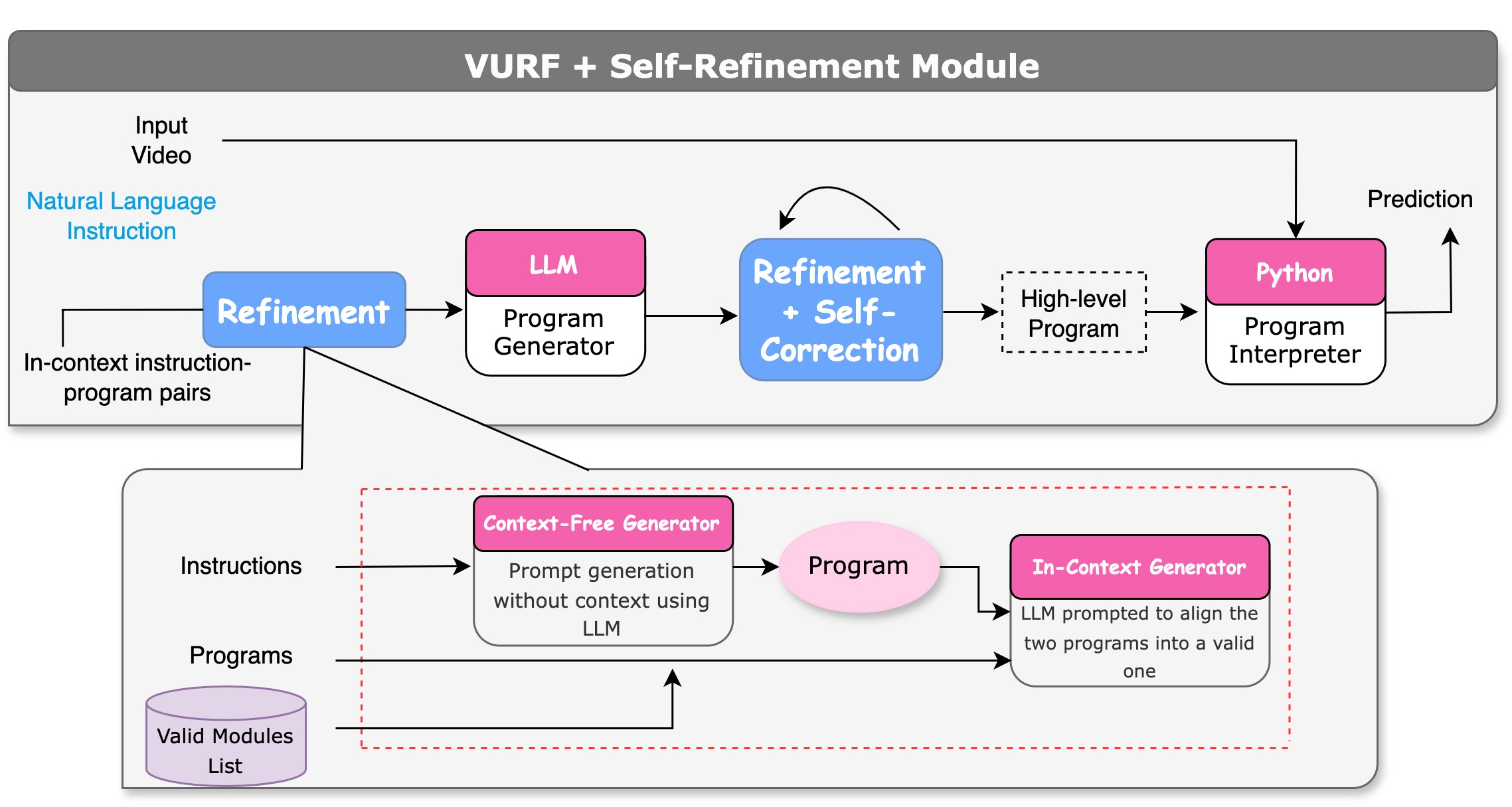}
  \caption{\textbf{Video Understanding and Reasoning Framework (VURF) pipeline.} \textbf{Top:} figure shows the main approach of VURF with the added self-correction module. \textbf{Bottom:} figure shows the self-refinement module.}
  \label{fig:refine}
\end{figure}

\section{Methodology}
\label{sec:method}

In contrast to conventional task-specific models that exhibit limitations in addressing complex reasoning challenges, the Video Understanding and Reasoning Framework (VURF) seeks to utilize the reasoning power of LLMs to deconstruct complex video-related queries into a series of sub-tasks (video programs). By executing these sub-tasks sequentially, we can culminate to arrive at the final response. Moreover, VURF allows seamless integration of new visual models in a plug-and-play manner and also employs self-critique mechanisms to mitigate LLM's hallucinations and judgment errors. An overview of our approach is shown in Fig.~\ref{fig:refine}, and we explain the approach in detail below.

\subsection{Video Reasoning LLMs}
\textbf{Role of Large Language Models.}
Large language models (LLMs), exemplified by GPT-3 \cite{brown2020language} and GPT-4 \cite{achiam2023gpt}, have demonstrated an impressive in-context learning capability (ICL) that does not necessitate fine-tuning. ICL aims to extend the understanding of LLMs to novel scenarios using a restricted set of input and output demonstrations within the relevant context. In this work, we leverage GPT-3.5 to generate visual programs that solve video reasoning tasks involving natural language instructions. The LLMs efficiently perform video reasoning tasks by avoiding direct video processing. Instead, it formulates logical sequence flows that decompose complex tasks into simpler sub-tasks. This approach enhances efficiency and allows the model to navigate the intricacies of video-related challenges.

\noindent\textbf{Prompting.}
We prompt GPT-3.5 with in-context examples which consist of pairs of instruction and the associated programs that the LLM is expected to generate. The programs follow a generic structure 
where each line of the program includes the name of a module, the module's input argument names and their values, and an output variable name. 
As output variables in a specific step are used later for another step they follow a general structure which the LLM learns:
\begin{lstlisting}[basicstyle=\ttfamily]
    OUTPUT0=FUNC0(video=VIDEO,...)
    OUTPUT1=FUNC1(arg0=OUTPUT0,...)
    ...
\end{lstlisting}
Given a set of these pairs and a new instruction, the LLM can generate a new program that follows the same structure and can thus be executed via our program interpreter.

\subsection{Self-Refinement}
One prominent limitation of a naive visual programming technique like  \cite{gupta2023visual}  is the proneness of generating inaccurate information influenced by contextual cues, as well as lacking an inherent capacity for autonomous self-correction through task-agnostic knowledge. 
Two major issues arise with the LLM-generated program. Firstly, due to LLM hallucinations, the program might be using some function that is not supported by our interpreter. Secondly, the program may not break down the instruction into sub-tasks in an optimal fashion. We resolve these limitations by the below steps.

\textbf{Error Correction:} To address the issue of a program utilizing a function unsupported by our interpreter, we employ a feedback generation approach, which notably leverages the power of GPT-3.5. We present the program to this module, alongside the available list of functions and their general usage, and inquire if the given program violates these constraints. If discrepancies are identified, the program is subsequently regenerated, incorporating the provided feedback as contextual information. This iterative process ensures error-free execution of the given instruction.

\begin{figure}[t!]  
  \includegraphics[width=\textwidth]{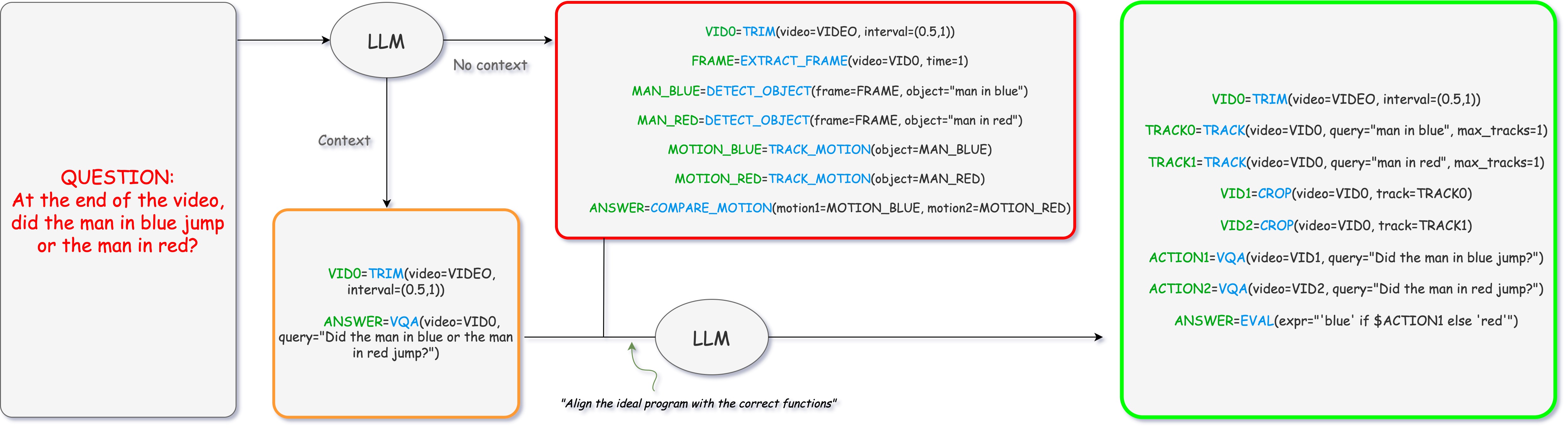}

  \caption{\textbf{Auto Self-Refinement} example. Two programs are generated: one with contextual examples and one without, but with added information for structural integrity. Both are then input into the Language Model (LLM) to generate a new program that aligns with the ideal while avoiding invalid functions.}
  \vspace{-1.75em}
  \label{fig:refine_eg}
\end{figure}

\textbf{Auto-Refinement of In-Context Examples:} The effectiveness of our approach hinges significantly on the quality of in-context examples. Therefore, it is imperative to refine these examples to enhance the module's performance \cite{lu2023self,madaan2023self}. To accomplish this, we implement a self-refinement procedure. Initially, given an instruction, $I$, and the initial program generated $P$, we input $I$ to the LLM, prompting it to generate an improved program, $P'$, without the inclusion of in-context examples. Subsequently, both $P$ and $P'$ are input into the LLM, enabling the generation of a refined program that aligns more closely with the LLM's reasoning, all while excluding the influence of in-context examples. Replicating this process for $n$ instructions yields a set of $n$ new in-context examples, enhancing the module's ability to perform tasks effectively. Note that as shown in Fig.~\ref{fig:refine}, the auto self-refinement module can be applied to a single user query to iteratively improve the generated program as well. However, this is inefficient and costly and thus we show that even pre-refining the in-context program-instruction pairs can improve the performance of the VURF (Table \ref{tab:Accuracies}). A concrete illustration of this process on a single program and instruction pair is provided in Fig. \ref{fig:refine_eg}.

\begin{figure}[t!]
    \includegraphics[width=\textwidth]{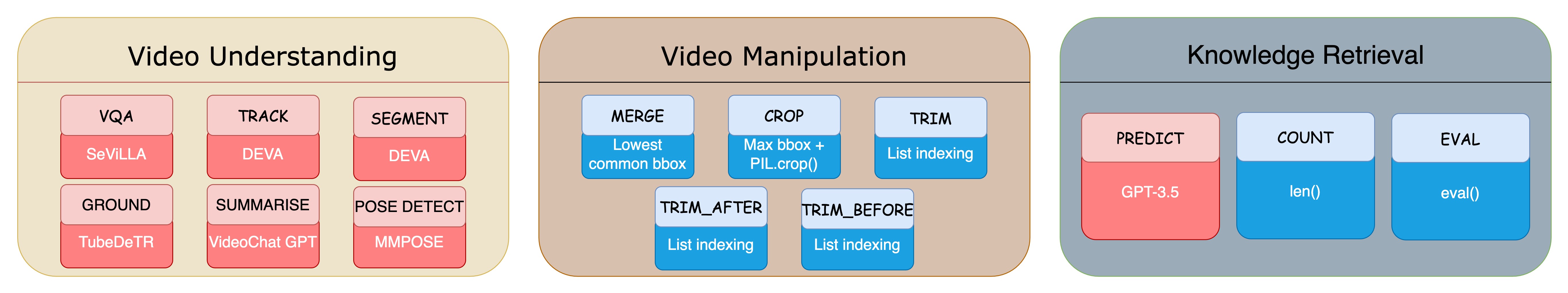}
    \caption{\textbf{Main Modules used by VURF.} The red boxes show modules that require a pre-trained model whereas the boxes are modules that require trivial functions.}
    \label{fig: modules}
\end{figure}

\section{Tasks}
\label{sec:Tasks}
Our video understanding and reasoning framework (VURF) aims to offer a versatile approach adaptable to various visual tasks. By integrating an interpreter component into an existing state-of-the-art (SOTA) vision model, VURF addresses four diverse challenges: Video Question Answering (VQA), Video Anticipation, Pose Estimation, and Multi-Video VQA. VURF functions by employing a large language model (LLM) as a reasoning module to generate a visual program. This program outlines a sequence of steps, each executed independently. The output of one step is fed as input to the next, creating a cohesive workflow.

\subsection{Video Question Answering}
\begin{figure}[t!]
    \includegraphics[width=\textwidth]{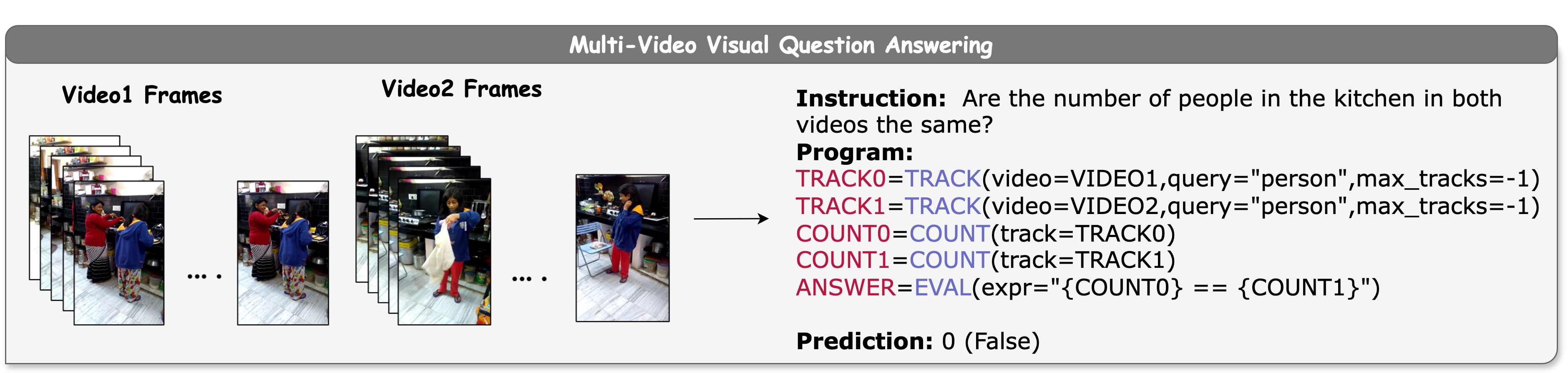}
    \caption{A qualitative example showing the Program steps in the \textbf{Multi-Video VQA task}. The programs provide a logical decomposition of the original complex tasks.}
    \label{fig:compVQA}
\end{figure}
VQA is an important task in video comprehension that endeavors to connect natural language processing with video comprehension. The objective is to empower models to interpret the content of a video and respond to user queries related to the video content. Effective resolution of such queries often involves breaking down the overarching problem into manageable sub-problems. 

For instance, consider a video of a man entering a room and then performing some action. If a human is asked, ``What does the man do after entering the room?'' they would visually inspect the video, locate the man, observe his actions by examining the video, and then deduce what the nature of his activities. This example illustrates the critical role of logical reasoning in decomposing complex questions into more manageable sub-queries, suitable for evaluation using off-the-shelf models. The inherent complexity of such tasks makes VQA an ideal candidate for a visual programming approach, given its primary function of decomposing intricate tasks into more manageable components. 
In the context of the aforementioned example, VURF, as demonstrated in Fig. \ref{fig:egs} (Right), initially employs a GROUNDING model to identify an interval where the man enters the room. Subsequently, it would call the TRIMAFTER module to retrieve the relevant part of the video, and finally using the VQA module, it will engage in visual examination of the video to answer the question ``Pick up towel''. This approach makes the task more logical and interpretable, with possible explanations in case an output is wrong.

VURF also extends its functionality to Multi-Video Question Answering tasks which involves synthesizing information from two distinct videos to provide accurate responses to user queries. Our approach simplifies this intricate process by breaking it down into a sequence of steps executed by a language model. An example is shown in Fig. \ref{fig:compVQA}.

\begin{figure*}[t]  
  \centering
  \includegraphics[width=\textwidth]{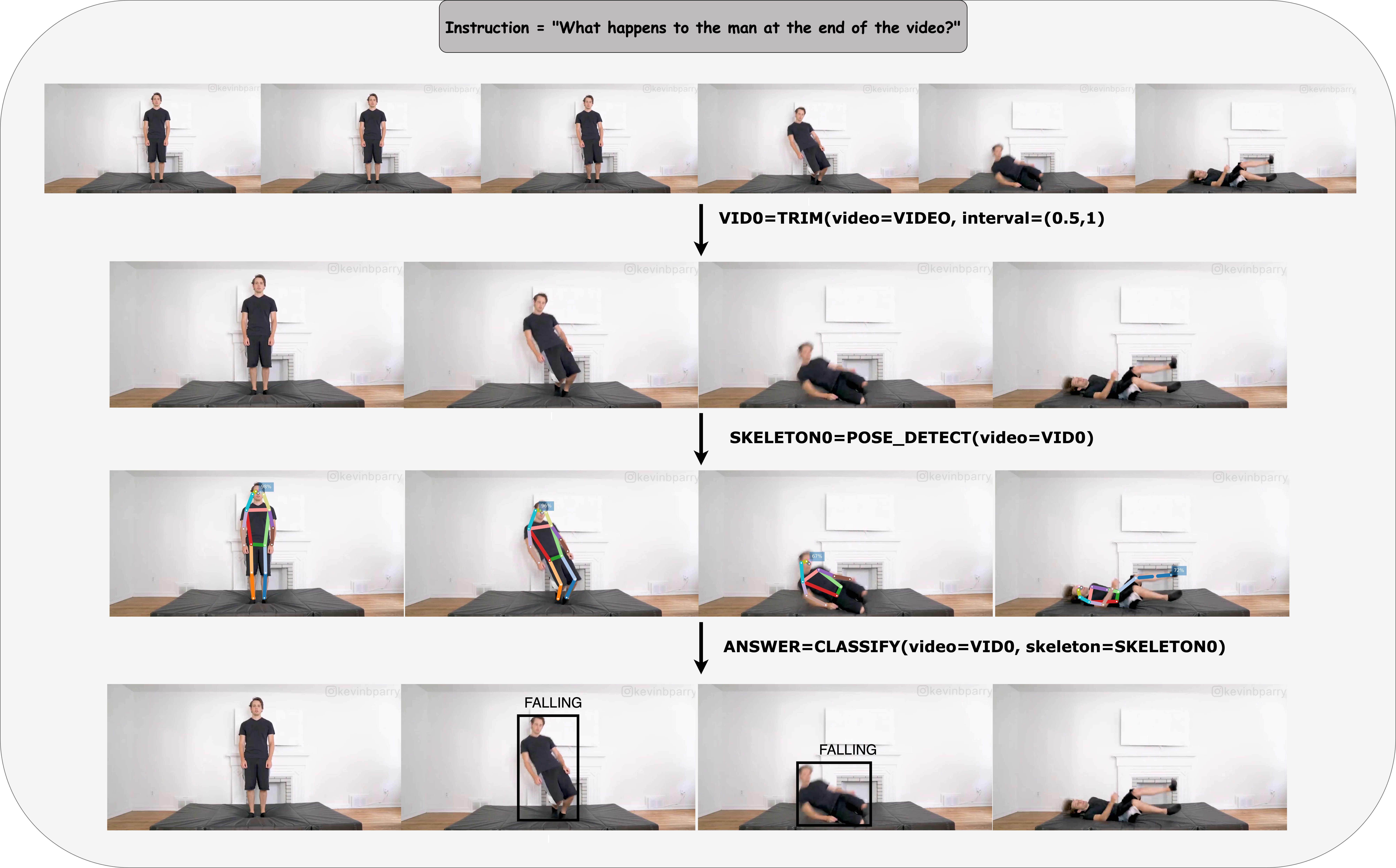}
  \caption{\textbf{Qualitative example} showing the program steps of the Pose Estimation task of VURF.}
  \label{fig:pose_est}
  \vspace{-1em}
\end{figure*}

\subsection{Pose Estimation}
Pose Estimation identifies the position and orientation of human bodies in images or videos using a skeletal model and has been applied in areas like HCI, virtual reality, and clinical assessments \cite{zheng2023deep, erol2007vision, escobar2019hand}. The process typically involves two stages: detecting joint orientations, as seen in tools like MMPose \cite{sengupta2020mm} and OpenPose \cite{cao2017realtime}, followed by using a model to estimate the pose from the skeletal representation \cite{andriluka20142d}. While these approaches are effective, they often struggle to generalize to unseen datasets.

In our Visual Programming approach, pose estimation is used to pre-process videos by tracking and cropping specific individuals, followed by pose classification. We utilize MMPose for keypoint detection and classify poses for tasks like fall detection, which can be extended to applications such as hazard or crime detection. For example, Fig. \ref{fig:egs} shows pose tagging in a video, and Fig. \ref{fig:pose_est} illustrates a fall detection scenario where relevant frames are trimmed, pose is detected, and the fall is classified.

\subsection{Video Editing}
Video editing is a crucial process in the post-production phase of film-making, television production, and other visual media industries \cite{dancyger2018technique}. It involves manipulating and rearranging video clips to create a coherent and engaging narrative or visual presentation. Video editing encompasses various tasks such as trimming, cutting, merging, and arranging video segments, as well as adding visual effects, transitions, and audio enhancements. In today's digital age, video editing is typically performed using specialized software applications that offer a wide range of tools and features to facilitate the editing process.

\begin{figure}[t!]  
  \includegraphics[width=\textwidth]{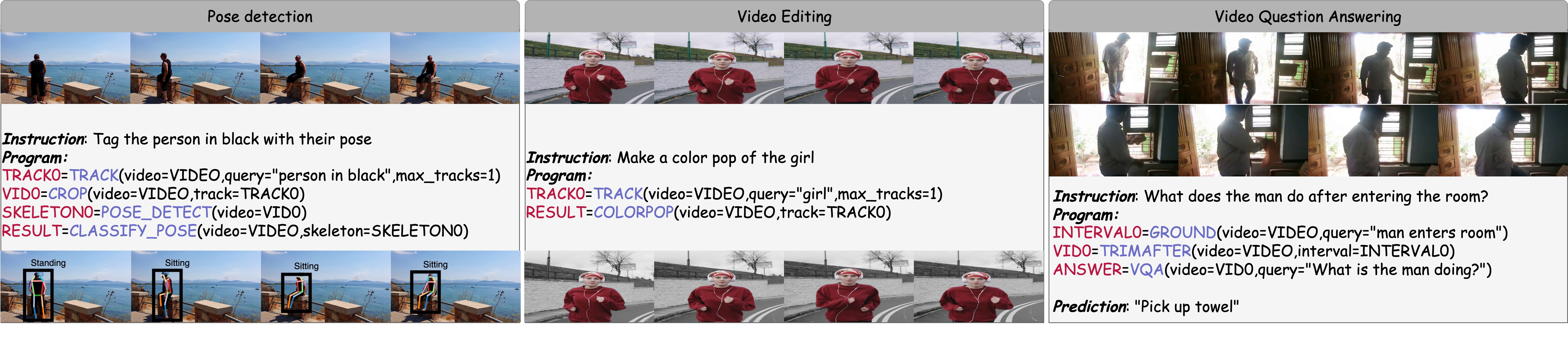}
  \caption{\textbf{Qualitative examples} demonstrating different use cases of VURF. \textbf{Left}: An example of Pose Detection that uses the visual program generated by VURF. \textbf{Middle}: A qualitative example of Video Editing that directly leverages VURF's list of modules. \textbf{Right}: An example of VQA task decomposition by manageable sub-tasks by our framework.
  \textit{Best viewed in zoom.}}
  \label{fig:egs}
\end{figure}

Our approach directly applies to Video Editing by breaking any instruction down into a series of sequential steps. For this task we employ the use of the MERGE, CROP, TRIM, BGBLUR, and COLORPOP functions. Some concrete qualitative examples can be found in Fig. \ref{fig:teaser} and \ref{fig:egs} (Middle).

\section{Experiments and Results}
\label{sec:Experiments and Results}
Our experiments encompass both quantitative and qualitative evaluation of the proposed video reasoning framework. 
In the quantitative experiments, we assess the performance impact of the proposed video programming approach built on a pre-trained model designed specifically for the Video Question Answering task. Additionally, we examine how the self-refinement approach affects the performance of the video programming approach, exploring the effects of altering the number of iterations on the model outputs. Finally, we present qualitative examples involving diverse tasks such as Pose Detection and Video Editing to highlight the efficacy of video programming in the context of video understanding tasks.

\subsection{VQA Evaluation}
Diverse approaches to video question answering (VQA) have been investigated, yielding promising outcomes. However, the SeViLA framework \cite{yu2024self}, has emerged as a distinguished approach, integrating temporal keyframe localization and question answering by employing a unified image-language model (BLIP2)\cite{li2023blip2}. We incorporate SeViLA as the VQA module, showcasing improved performance in the zero-shot video question-answering task with our VURF. The assessment is conducted on four benchmark datasets: STAR, NextQA, Social-IQ QA, and TrafficQA \cite{xu2021sutdtrafficqa}.

We conduct a comprehensive evaluation, commencing with the application of the base SeViLA zero-shot model on each dataset. Subsequently, we assess the effectiveness of the video programming approach on the same validation set. For the latter, a curated set of in-context examples is manually assembled for ICL, tailored to the specific characteristics of each dataset.

Several approaches, including InternVideo \cite{wang2022internvideo} and ViperGPT \cite{surís2023vipergpt}, have been proposed to address the challenge of Visual Question Answering alongside SeViLA. However, as illustrated in Table \ref{tab:Accuracies}, VURF outperforms all three in zero-shot video question answering. VURF offers a significant advantage over SeViLA and InternVideo by enabling reasoning through instructions rather than functioning as a black box, facilitating self-improvement. This capability proves advantageous over ViperGPT, which, despite employing the reasoning power of LLMs, is outperformed by VURF. The primary reason for this discrepancy is ViperGPT's susceptibility to contextual hallucinations. We demonstrate that integrating the self-refinement of in-context program-instruction pairs and incorporating a simple error correction module substantially enhances the performance of a visual programming approach.

\subsection{VQA Ablations}

ViperGPT \cite{surís2023vipergpt} diverges from VURF by not incorporating self-refinement mechanisms. This distinction is noteworthy as it highlights differing approaches to handling instruction comprehension within the context of VQA. Sole reliance on pre-trained language models without iterative refinement mechanisms makes the approach susceptible to contextual hallucinations. Our ablations (see Table \ref{tab: ablation}) clearly demonstrate that integrating the self-refinement of in-context program-instruction pairs and incorporating a simple error correction module substantially enhances the performance of a visual programming approach. 

Moreover, we also compare our method with visual instruction tuning \cite{liu2023visualinstructiontuning}. To this end, we finetune a LlaVA-based model (trained using Visual instruction tuning) called Video-ChatGPT \cite{Maaz2023VideoChatGPT} on question-program pairs and test the performance of generated programs on the STAR VQA dataset. VURF outperforms the visual instruction tuning method (Table \ref{tab: tune}), even when the latter has access to more data during the training phase compared to the number of in-context examples used by VURF. VURF's superior performance stems from its dynamic adaptation to various scenarios, leveraging contextual cues to manage tasks on the fly.

\begin{table}[!t]
    \caption{\emph{\textbf{Performance of VURF on Video Question Answering (zero-shot) compared to other existing models.}} Each result is evaluated on the validation set of the corresponding dataset.}
    \label{tab:Accuracies}
    
    \centering\small
    \renewcommand{\arraystretch}{1} 
    \setlength{\tabcolsep}{9.3pt} 
    \scalebox{0.85}[0.85]{
        \begin{tabular}{cccccc}
            \toprule
            \textbf{Datasets} & NextQA & STAR & Social-IQ-2.0 & TrafficQA \\
            \toprule
            \textbf{InternVideo} \cite{wang2022internvideo}  & 50.2\% & 41.8\% & 30.1\% & 31.2\% \\
            \textbf{ViperGPT} \cite{surís2023vipergpt} & 60.0\% & 40.3\% & 37.8\% & 35.7\% \\
            \midrule
            \textbf{SeViLA} \cite{yu2024self} & 63.8\% & 44.3\% & 47.3\% & 39.1\% \\
            \textbf{VURF} & \textcolor{ForestGreen}{64.0\%} & \textcolor{ForestGreen}{47.2\%} & \textcolor{ForestGreen}{51.6\%} & \textcolor{ForestGreen}{43.5\%} \\
            \bottomrule
        \end{tabular}
    }
\end{table}
\begin{table}[!t]
        \caption{\emph{\textbf{Comparison with Visual Instruction Tuning.}} We finetune Video-ChatGPT on program-question pairs (which we use as in-context examples in our system) and then test their program-generating performance.}
        \label{tab: tune}
	\centering\small
            \setlength{\tabcolsep}{9.3pt} 
            \renewcommand{\arraystretch}{1} 
        \scalebox{0.85}[0.85]{
       \begin{tabular}{ccc}
            \toprule
    	\textbf{Dataset} & VURF (ours) & Visual Instruction Tuning \\
            \toprule
            \textbf{STAR} & \textcolor{ForestGreen}{$44.5\%$} & $22.3\%$ \\
    	\bottomrule
    \end{tabular}}
\end{table}

\subsection{Self-Refinement}
\begin{enumerate}
     \item \textbf{In-Context Example Refinement:} In examining the self-refinement process, we employ the NeXt-QA dataset by randomly sampling 50 videos for the test set. Additionally, we curate a set of 20 in-context examples for evaluation. The accuracy of the test set is initially calculated. Subsequently, these in-context examples undergo self-refinement. This approach initially presents instructions to an LLM for program generation without contextual influence. The generated program, along with the original program, is then provided to the LLM with a prompt to maintain the structure of the initial program while enhancing it using the non-contextual program. The refined in-context examples undergo multiple iterations of this process, and we report the accuracy on the test set concerning the number of iterations of self-refinement applied to the in-context examples in Fig. \ref{fig: acc_vs_iter}.

    \item \textbf{Error Correction Evaluation:} To assess the efficacy of automatic error correction, we randomly select 400 videos from the STAR dataset. The evaluation involves calculating the number of errors stemming from program invalidity. Given the ability to feed a program into the auto-correction module multiple times, we explore the impact of increasing these iterations on error occurrences within the 400 video-instruction pairs and report our evaluations in Fig. \ref{fig: error_vs_iter}.

    
    \begin{figure}[t!]
        \begin{subfigure}{0.48\textwidth}
            \centering
            \includegraphics[width=0.92\linewidth]{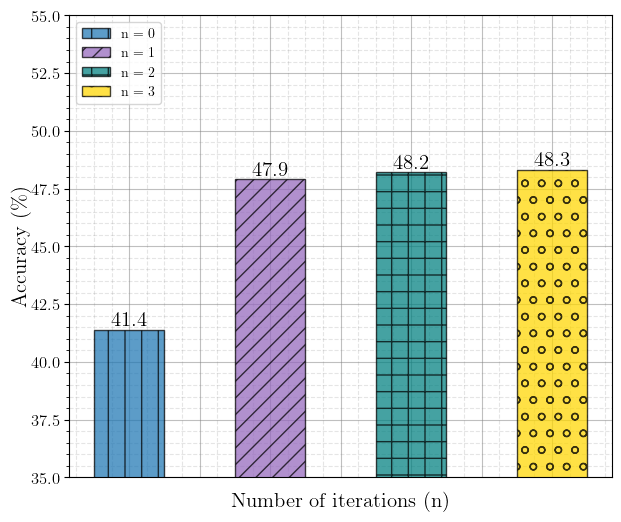}
            \caption{\textbf{Accuracy plotted against the number of iterations of self-refinement}.}
            \label{fig: acc_vs_iter}
        \end{subfigure}\hfill
        \begin{subfigure}{0.48\textwidth}
            \centering
            \includegraphics[width=\linewidth]{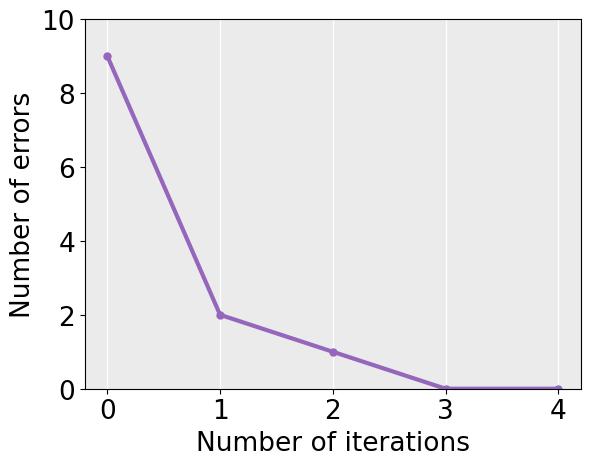}
            \caption{\textbf{Number of errors plotted against the number of iterations of self-refinement.}}
            \label{fig: error_vs_iter}
        \end{subfigure}
    \caption{The impact of the self-refinement stage on quality of video programs. }
    \label{fig: combined}
\end{figure}
    
\begin{table}[!t]
    \caption{\emph{\textbf{Ablations demonstrating the effectiveness of the refinement pipeline.}} Results that do not use the error correction module, are such that if a syntactically incorrect program is generated it is considered as a wrong prediction.}
	\label{tab: ablation}
 
	\centering\small
		\scalebox{0.85}[0.85]{
		\begin{tabular}{lcccc}
                    \toprule
				\textbf{Datasets} & NextQA & STAR & Social-IQ-2.0 & TrafficQA \\
                    \midrule
                    Accuracy \textbf{\textit{w/o error correction \& self refinement}}& 47.8\% & 42.1\% & 45.5\% & 38.3\% \\
                    Accuracy \textbf{\textit{w/o error correction}} & 57.1\% & 44.9\% & 48.3\% & 41.1\% \\
                    Accuracy \textbf{\textit{w/o self refinement}}& 56.5\% & 43.1\% & 47.2\% & 40.2\% \\
                    \toprule
                    Accuracy \textbf{\textit{with self refinement \& error correction}} & \textcolor{ForestGreen}{64.0\%} &  \textcolor{ForestGreen}{47.2\%} & \textcolor{ForestGreen}{51.6\%} & \textcolor{ForestGreen}{43.5\%} \\
	   \bottomrule
	\end{tabular}}
	
\end{table}
\end{enumerate}

\section{Conclusion}
\label{sec:Conclusion}
Our work expands the boundaries of Visual Programming by integrating it into the realm of video understanding, showcasing its efficacy in diverse applications such as Video Question Answering, Pose Estimation, and Multi-Video VQA. Additionally, we introduce a novel approach for enhancing the program generation process within the LLM through self-refinement, consequently elevating the efficacy of few-shot prompting to the LLM. Our approach not only broadens the scope of Visual Programming but also underscores the potential for continuous self-refinement to optimize the capabilities of LLMs for video reasoning tasks.

%
%
\bibliographystyle{plainnat}
\bibliography{main}

\begin{thebibliography}{70}
\providecommand{\natexlab}[1]{#1}
\providecommand{\url}[1]{\texttt{#1}}
\expandafter\ifx\csname urlstyle\endcsname\relax
  \providecommand{\doi}[1]{doi: #1}\else
  \providecommand{\doi}{doi: \begingroup \urlstyle{rm}\Url}\fi

\bibitem[Achiam et~al.(2023)Achiam, Adler, Agarwal, Ahmad, Akkaya, Aleman, Almeida, Altenschmidt, Altman, Anadkat, et~al.]{achiam2023gpt}
Josh Achiam, Steven Adler, Sandhini Agarwal, Lama Ahmad, Ilge Akkaya, Florencia~Leoni Aleman, Diogo Almeida, Janko Altenschmidt, Sam Altman, Shyamal Anadkat, et~al.
\newblock Gpt-4 technical report.
\newblock \emph{arXiv preprint arXiv:2303.08774}, 2023.

\bibitem[Alayrac et~al.(2022)Alayrac, Donahue, Luc, Miech, Barr, Hasson, Lenc, Mensch, Millican, Reynolds, et~al.]{alayrac2022flamingo}
Jean-Baptiste Alayrac, Jeff Donahue, Pauline Luc, Antoine Miech, Iain Barr, Yana Hasson, Karel Lenc, Arthur Mensch, Katherine Millican, Malcolm Reynolds, et~al.
\newblock Flamingo: a visual language model for few-shot learning.
\newblock \emph{Advances in Neural Information Processing Systems}, 35:\penalty0 23716--23736, 2022.

\bibitem[Andriluka et~al.(2014)Andriluka, Pishchulin, Gehler, and Schiele]{andriluka20142d}
Mykhaylo Andriluka, Leonid Pishchulin, Peter Gehler, and Bernt Schiele.
\newblock 2d human pose estimation: New benchmark and state of the art analysis.
\newblock In \emph{Proceedings of the IEEE Conference on computer Vision and Pattern Recognition}, pages 3686--3693, 2014.

\bibitem[Antol et~al.(2015)Antol, Agrawal, Lu, Mitchell, Batra, Zitnick, and Parikh]{antol2015vqa}
Stanislaw Antol, Aishwarya Agrawal, Jiasen Lu, Margaret Mitchell, Dhruv Batra, C~Lawrence Zitnick, and Devi Parikh.
\newblock Vqa: Visual question answering.
\newblock In \emph{Proceedings of the IEEE international conference on computer vision}, pages 2425--2433, 2015.

\bibitem[Arnab et~al.(2021)Arnab, Dehghani, Heigold, Sun, Lu{\v{c}}i{\'c}, and Schmid]{arnab2021vivit}
Anurag Arnab, Mostafa Dehghani, Georg Heigold, Chen Sun, Mario Lu{\v{c}}i{\'c}, and Cordelia Schmid.
\newblock Vivit: A video vision transformer.
\newblock In \emph{Proceedings of the IEEE/CVF international conference on computer vision}, pages 6836--6846, 2021.

\bibitem[Brown et~al.(2020)Brown, Mann, Ryder, Subbiah, Kaplan, Dhariwal, Neelakantan, Shyam, Sastry, Askell, et~al.]{brown2020language}
Tom Brown, Benjamin Mann, Nick Ryder, Melanie Subbiah, Jared~D Kaplan, Prafulla Dhariwal, Arvind Neelakantan, Pranav Shyam, Girish Sastry, Amanda Askell, et~al.
\newblock Language models are few-shot learners.
\newblock \emph{Advances in neural information processing systems}, 33:\penalty0 1877--1901, 2020.

\bibitem[Campos and Shern(2022)]{campos2022training}
Jon~Ander Campos and Jun Shern.
\newblock Training language models with language feedback.
\newblock In \emph{ACL Workshop on Learning with Natural Language Supervision. 2022.}, 2022.

\bibitem[Cao et~al.(2017)Cao, Simon, Wei, and Sheikh]{cao2017realtime}
Zhe Cao, Tomas Simon, Shih-En Wei, and Yaser Sheikh.
\newblock Realtime multi-person 2d pose estimation using part affinity fields.
\newblock In \emph{Proceedings of the IEEE conference on computer vision and pattern recognition}, pages 7291--7299, 2017.

\bibitem[Carreira and Zisserman(2017)]{carreira2017quo}
Joao Carreira and Andrew Zisserman.
\newblock Quo vadis, action recognition? a new model and the kinetics dataset.
\newblock In \emph{proceedings of the IEEE Conference on Computer Vision and Pattern Recognition}, pages 6299--6308, 2017.

\bibitem[Chen et~al.(2018)Chen, Chen, Ma, Jie, and Chua]{chen2018temporally}
Jingyuan Chen, Xinpeng Chen, Lin Ma, Zequn Jie, and Tat-Seng Chua.
\newblock Temporally grounding natural sentence in video.
\newblock In \emph{Proceedings of the 2018 conference on empirical methods in natural language processing}, pages 162--171, 2018.

\bibitem[Chen et~al.(2019)Chen, Gan, Shen, Huang, Zeng, and Tan]{chen2019relation}
Peihao Chen, Chuang Gan, Guangyao Shen, Wenbing Huang, Runhao Zeng, and Mingkui Tan.
\newblock Relation attention for temporal action localization.
\newblock \emph{IEEE Transactions on Multimedia}, 22\penalty0 (10):\penalty0 2723--2733, 2019.

\bibitem[Chen et~al.(2023)Chen, Lin, Sch{\"a}rli, and Zhou]{chen2023teaching}
Xinyun Chen, Maxwell Lin, Nathanael Sch{\"a}rli, and Denny Zhou.
\newblock Teaching large language models to self-debug.
\newblock \emph{arXiv preprint arXiv:2304.05128}, 2023.

\bibitem[Dancyger(2018)]{dancyger2018technique}
Ken Dancyger.
\newblock \emph{The technique of film and video editing: history, theory, and practice}.
\newblock Routledge, 2018.

\bibitem[Deliege et~al.(2021)Deliege, Cioppa, Giancola, Seikavandi, Dueholm, Nasrollahi, Ghanem, Moeslund, and Van~Droogenbroeck]{deliege2021soccernet}
Adrien Deliege, Anthony Cioppa, Silvio Giancola, Meisam~J Seikavandi, Jacob~V Dueholm, Kamal Nasrollahi, Bernard Ghanem, Thomas~B Moeslund, and Marc Van~Droogenbroeck.
\newblock Soccernet-v2: A dataset and benchmarks for holistic understanding of broadcast soccer videos.
\newblock In \emph{Proceedings of the IEEE/CVF conference on computer vision and pattern recognition}, pages 4508--4519, 2021.

\bibitem[Erol et~al.(2007)Erol, Bebis, Nicolescu, Boyle, and Twombly]{erol2007vision}
Ali Erol, George Bebis, Mircea Nicolescu, Richard~D Boyle, and Xander Twombly.
\newblock Vision-based hand pose estimation: A review.
\newblock \emph{Computer Vision and Image Understanding}, 108\penalty0 (1-2):\penalty0 52--73, 2007.

\bibitem[Escobar et~al.(2019)Escobar, Gonz{\'a}lez, Torres, Daza, Triana, and Arbel{\'a}ez]{escobar2019hand}
Mar{\'\i}a Escobar, Cristina Gonz{\'a}lez, Felipe Torres, Laura Daza, Gustavo Triana, and Pablo Arbel{\'a}ez.
\newblock Hand pose estimation for pediatric bone age assessment.
\newblock In \emph{Medical Image Computing and Computer Assisted Intervention--MICCAI 2019: 22nd International Conference, Shenzhen, China, October 13--17, 2019, Proceedings, Part VI 22}, pages 531--539. Springer, 2019.

\bibitem[Evans et~al.(2021)Evans, Cotton-Barratt, Finnveden, Bales, Balwit, Wills, Righetti, and Saunders]{evans2021truthful}
Owain Evans, Owen Cotton-Barratt, Lukas Finnveden, Adam Bales, Avital Balwit, Peter Wills, Luca Righetti, and William Saunders.
\newblock Truthful ai: Developing and governing ai that does not lie.
\newblock \emph{arXiv preprint arXiv:2110.06674}, 2021.

\bibitem[Feng et~al.(2024)Feng, Zhang, Li, Liu, Lang, Feng, Wu, and Liu]{feng2024improving}
Zhaopeng Feng, Yan Zhang, Hao Li, Wenqiang Liu, Jun Lang, Yang Feng, Jian Wu, and Zuozhu Liu.
\newblock Improving llm-based machine translation with systematic self-correction.
\newblock \emph{arXiv preprint arXiv:2402.16379}, 2024.

\bibitem[Gammulle et~al.(2019)Gammulle, Denman, Sridharan, and Fookes]{gammulle2019predicting}
Harshala Gammulle, Simon Denman, Sridha Sridharan, and Clinton Fookes.
\newblock Predicting the future: A jointly learnt model for action anticipation.
\newblock In \emph{Proceedings of the IEEE/CVF International Conference on Computer Vision}, pages 5562--5571, 2019.

\bibitem[Gao et~al.(2017)Gao, Sun, Yang, and Nevatia]{gao2017tall}
Jiyang Gao, Chen Sun, Zhenheng Yang, and Ram Nevatia.
\newblock Tall: Temporal activity localization via language query.
\newblock In \emph{Proceedings of the IEEE international conference on computer vision}, pages 5267--5275, 2017.

\bibitem[Girdhar and Grauman(2021)]{girdhar2021anticipative}
Rohit Girdhar and Kristen Grauman.
\newblock Anticipative video transformer.
\newblock In \emph{Proceedings of the IEEE/CVF international conference on computer vision}, pages 13505--13515, 2021.

\bibitem[Golovneva et~al.(2022)Golovneva, Chen, Poff, Corredor, Zettlemoyer, Fazel-Zarandi, and Celikyilmaz]{golovneva2022roscoe}
Olga Golovneva, Moya Chen, Spencer Poff, Martin Corredor, Luke Zettlemoyer, Maryam Fazel-Zarandi, and Asli Celikyilmaz.
\newblock Roscoe: A suite of metrics for scoring step-by-step reasoning.
\newblock \emph{arXiv preprint arXiv:2212.07919}, 2022.

\bibitem[Gou et~al.(2023)Gou, Shao, Gong, Shen, Yang, Duan, and Chen]{gou2023critic}
Zhibin Gou, Zhihong Shao, Yeyun Gong, Yelong Shen, Yujiu Yang, Nan Duan, and Weizhu Chen.
\newblock Critic: Large language models can self-correct with tool-interactive critiquing.
\newblock \emph{arXiv preprint arXiv:2305.11738}, 2023.

\bibitem[Gu et~al.(2018)Gu, Sun, Ross, Vondrick, Pantofaru, Li, Vijayanarasimhan, Toderici, Ricco, Sukthankar, et~al.]{gu2018ava}
Chunhui Gu, Chen Sun, David~A Ross, Carl Vondrick, Caroline Pantofaru, Yeqing Li, Sudheendra Vijayanarasimhan, George Toderici, Susanna Ricco, Rahul Sukthankar, et~al.
\newblock Ava: A video dataset of spatio-temporally localized atomic visual actions.
\newblock In \emph{Proceedings of the IEEE conference on computer vision and pattern recognition}, pages 6047--6056, 2018.

\bibitem[Gupta and Kembhavi(2023)]{gupta2023visual}
Tanmay Gupta and Aniruddha Kembhavi.
\newblock Visual programming: Compositional visual reasoning without training.
\newblock In \emph{Proceedings of the IEEE/CVF Conference on Computer Vision and Pattern Recognition}, pages 14953--14962, 2023.

\bibitem[Huang et~al.(2020)Huang, Xiong, Rao, Wang, and Lin]{huang2020movienet}
Qingqiu Huang, Yu~Xiong, Anyi Rao, Jiaze Wang, and Dahua Lin.
\newblock Movienet: A holistic dataset for movie understanding.
\newblock In \emph{Computer Vision--ECCV 2020: 16th European Conference, Glasgow, UK, August 23--28, 2020, Proceedings, Part IV 16}, pages 709--727. Springer, 2020.

\bibitem[Kim et~al.(2023)Kim, Baldi, and McAleer]{kim2023language}
Geunwoo Kim, Pierre Baldi, and Stephen McAleer.
\newblock Language models can solve computer tasks.
\newblock \emph{arXiv preprint arXiv:2303.17491}, 2023.

\bibitem[Koprinska and Carrato(2001)]{koprinska2001temporal}
Irena Koprinska and Sergio Carrato.
\newblock Temporal video segmentation: A survey.
\newblock \emph{Signal processing: Image communication}, 16\penalty0 (5):\penalty0 477--500, 2001.

\bibitem[Le et~al.(2022)Le, Wang, Gotmare, Savarese, and Hoi]{le2022coderl}
Hung Le, Yue Wang, Akhilesh~Deepak Gotmare, Silvio Savarese, and Steven Chu~Hong Hoi.
\newblock Coderl: Mastering code generation through pretrained models and deep reinforcement learning.
\newblock \emph{Advances in Neural Information Processing Systems}, 35:\penalty0 21314--21328, 2022.

\bibitem[Lei et~al.(2018)Lei, Yu, Bansal, and Berg]{lei2018tvqa}
Jie Lei, Licheng Yu, Mohit Bansal, and Tamara~L Berg.
\newblock Tvqa: Localized, compositional video question answering.
\newblock \emph{arXiv preprint arXiv:1809.01696}, 2018.

\bibitem[Li et~al.(2023)Li, Li, Savarese, and Hoi]{li2023blip2}
Junnan Li, Dongxu Li, Silvio Savarese, and Steven Hoi.
\newblock Blip-2: Bootstrapping language-image pre-training with frozen image encoders and large language models, 2023.

\bibitem[Li et~al.(2020)Li, Gao, Niu, Xiao, Liu, Liu, Wu, and Wang]{li2020unimo}
Wei Li, Can Gao, Guocheng Niu, Xinyan Xiao, Hao Liu, Jiachen Liu, Hua Wu, and Haifeng Wang.
\newblock Unimo: Towards unified-modal understanding and generation via cross-modal contrastive learning.
\newblock \emph{arXiv preprint arXiv:2012.15409}, 2020.

\bibitem[Li et~al.(2022)Li, Wu, Fan, Mangalam, Xiong, Malik, and Feichtenhofer]{li2022mvitv2}
Yanghao Li, Chao-Yuan Wu, Haoqi Fan, Karttikeya Mangalam, Bo~Xiong, Jitendra Malik, and Christoph Feichtenhofer.
\newblock Mvitv2: Improved multiscale vision transformers for classification and detection.
\newblock In \emph{Proceedings of the IEEE/CVF Conference on Computer Vision and Pattern Recognition}, pages 4804--4814, 2022.

\bibitem[Lin et~al.(2023{\natexlab{a}})Lin, Zhang, Chen, Pramanick, Gao, Wang, Yan, and Shou]{lin2023univtg}
Kevin~Qinghong Lin, Pengchuan Zhang, Joya Chen, Shraman Pramanick, Difei Gao, Alex~Jinpeng Wang, Rui Yan, and Mike~Zheng Shou.
\newblock Univtg: Towards unified video-language temporal grounding.
\newblock In \emph{Proceedings of the IEEE/CVF International Conference on Computer Vision}, pages 2794--2804, 2023{\natexlab{a}}.

\bibitem[Lin et~al.(2023{\natexlab{b}})Lin, Tiwari, Huang, Li, Shou, Ji, and Chang]{lin2023towards}
Xudong Lin, Simran Tiwari, Shiyuan Huang, Manling Li, Mike~Zheng Shou, Heng Ji, and Shih-Fu Chang.
\newblock Towards fast adaptation of pretrained contrastive models for multi-channel video-language retrieval.
\newblock In \emph{Proceedings of the IEEE/CVF Conference on Computer Vision and Pattern Recognition}, pages 14846--14855, 2023{\natexlab{b}}.

\bibitem[Liu et~al.(2023)Liu, Li, Wu, and Lee]{liu2023visualinstructiontuning}
Haotian Liu, Chunyuan Li, Qingyang Wu, and Yong~Jae Lee.
\newblock Visual instruction tuning, 2023.
\newblock URL \url{https://arxiv.org/abs/2304.08485}.

\bibitem[Liu et~al.(2022)Liu, Ning, Cao, Wei, Zhang, Lin, and Hu]{liu2022video}
Ze~Liu, Jia Ning, Yue Cao, Yixuan Wei, Zheng Zhang, Stephen Lin, and Han Hu.
\newblock Video swin transformer.
\newblock In \emph{Proceedings of the IEEE/CVF conference on computer vision and pattern recognition}, pages 3202--3211, 2022.

\bibitem[Lu et~al.(2023)Lu, Zhong, Huang, Wang, Mi, Wang, Wang, Shang, and Liu]{lu2023self}
Jianqiao Lu, Wanjun Zhong, Wenyong Huang, Yufei Wang, Fei Mi, Baojun Wang, Weichao Wang, Lifeng Shang, and Qun Liu.
\newblock Self: Language-driven self-evolution for large language model.
\newblock \emph{arXiv preprint arXiv:2310.00533}, 2023.

\bibitem[Maaz et~al.(2024)Maaz, Rasheed, Khan, and Khan]{Maaz2023VideoChatGPT}
Muhammad Maaz, Hanoona Rasheed, Salman Khan, and Fahad~Shahbaz Khan.
\newblock Video-chatgpt: Towards detailed video understanding via large vision and language models.
\newblock In \emph{Proceedings of the 62nd Annual Meeting of the Association for Computational Linguistics (ACL 2024)}, 2024.

\bibitem[Madaan et~al.(2023)Madaan, Tandon, Gupta, Hallinan, Gao, Wiegreffe, Alon, Dziri, Prabhumoye, Yang, et~al.]{madaan2023self}
Aman Madaan, Niket Tandon, Prakhar Gupta, Skyler Hallinan, Luyu Gao, Sarah Wiegreffe, Uri Alon, Nouha Dziri, Shrimai Prabhumoye, Yiming Yang, et~al.
\newblock Self-refine: Iterative refinement with self-feedback.
\newblock \emph{arXiv preprint arXiv:2303.17651}, 2023.

\bibitem[Madaan et~al.(2024)Madaan, Tandon, Gupta, Hallinan, Gao, Wiegreffe, Alon, Dziri, Prabhumoye, Yang, et~al.]{madaan2024self}
Aman Madaan, Niket Tandon, Prakhar Gupta, Skyler Hallinan, Luyu Gao, Sarah Wiegreffe, Uri Alon, Nouha Dziri, Shrimai Prabhumoye, Yiming Yang, et~al.
\newblock Self-refine: Iterative refinement with self-feedback.
\newblock \emph{Advances in Neural Information Processing Systems}, 36, 2024.

\bibitem[Reddy et~al.(2023)Reddy, Shah, Paul, Mocharla, Hoffman, Katyal, Manocha, de~Melo, and Chellappa]{reddy2023synthetic}
Arun~V Reddy, Ketul Shah, William Paul, Rohita Mocharla, Judy Hoffman, Kapil~D Katyal, Dinesh Manocha, Celso~M de~Melo, and Rama Chellappa.
\newblock Synthetic-to-real domain adaptation for action recognition: A dataset and baseline performances.
\newblock \emph{arXiv preprint arXiv:2303.10280}, 2023.

\bibitem[Ruan et~al.(2023)Ruan, Chen, Zhang, Xu, Bao, Du, Shi, Mao, Zeng, and Zhao]{ruan2023tptu}
Jingqing Ruan, Yihong Chen, Bin Zhang, Zhiwei Xu, Tianpeng Bao, Guoqing Du, Shiwei Shi, Hangyu Mao, Xingyu Zeng, and Rui Zhao.
\newblock Tptu: Task planning and tool usage of large language model-based ai agents.
\newblock \emph{arXiv preprint arXiv:2308.03427}, 2023.

\bibitem[Sadhu et~al.(2021)Sadhu, Gupta, Yatskar, Nevatia, and Kembhavi]{sadhu2021visual}
Arka Sadhu, Tanmay Gupta, Mark Yatskar, Ram Nevatia, and Aniruddha Kembhavi.
\newblock Visual semantic role labeling for video understanding.
\newblock In \emph{Proceedings of the IEEE/CVF Conference on Computer Vision and Pattern Recognition}, pages 5589--5600, 2021.

\bibitem[Saunders et~al.(2022)Saunders, Yeh, Wu, Bills, Ouyang, Ward, and Leike]{saunders2022self}
William Saunders, Catherine Yeh, Jeff Wu, Steven Bills, Long Ouyang, Jonathan Ward, and Jan Leike.
\newblock Self-critiquing models for assisting human evaluators.
\newblock \emph{arXiv preprint arXiv:2206.05802}, 2022.

\bibitem[Sengupta et~al.(2020)Sengupta, Jin, Zhang, and Cao]{sengupta2020mm}
Arindam Sengupta, Feng Jin, Renyuan Zhang, and Siyang Cao.
\newblock mm-pose: Real-time human skeletal posture estimation using mmwave radars and cnns.
\newblock \emph{IEEE Sensors Journal}, 20\penalty0 (17):\penalty0 10032--10044, 2020.

\bibitem[Sigurdsson et~al.(2016)Sigurdsson, Varol, Wang, Farhadi, Laptev, and Gupta]{sigurdsson2016hollywood}
Gunnar~A Sigurdsson, G{\"u}l Varol, Xiaolong Wang, Ali Farhadi, Ivan Laptev, and Abhinav Gupta.
\newblock Hollywood in homes: Crowdsourcing data collection for activity understanding.
\newblock In \emph{Computer Vision--ECCV 2016: 14th European Conference, Amsterdam, The Netherlands, October 11--14, 2016, Proceedings, Part I 14}, pages 510--526. Springer, 2016.

\bibitem[Song et~al.(2023)Song, Chai, Wang, Zhang, Zhou, Wu, Guo, Ye, Lu, Hwang, et~al.]{song2023moviechat}
Enxin Song, Wenhao Chai, Guanhong Wang, Yucheng Zhang, Haoyang Zhou, Feiyang Wu, Xun Guo, Tian Ye, Yan Lu, Jenq-Neng Hwang, et~al.
\newblock Moviechat: From dense token to sparse memory for long video understanding.
\newblock \emph{arXiv preprint arXiv:2307.16449}, 2023.

\bibitem[Sun et~al.(2023)Sun, Xu, Zha, Liu, and Dong]{sun2023head}
Kai Sun, Yifan~Ethan Xu, Hanwen Zha, Yue Liu, and Xin~Luna Dong.
\newblock Head-to-tail: How knowledgeable are large language models (llm)? aka will llms replace knowledge graphs?
\newblock \emph{arXiv preprint arXiv:2308.10168}, 2023.

\bibitem[Surís et~al.(2023)Surís, Menon, and Vondrick]{surís2023vipergpt}
Dídac Surís, Sachit Menon, and Carl Vondrick.
\newblock Vipergpt: Visual inference via python execution for reasoning, 2023.

\bibitem[Thawakar et~al.(2024)Thawakar, Vayani, Khan, Cholakal, Anwer, Felsberg, Baldwin, Xing, and Khan]{thawakar2024mobillama}
Omkar Thawakar, Ashmal Vayani, Salman Khan, Hisham Cholakal, Rao~M Anwer, Michael Felsberg, Tim Baldwin, Eric~P Xing, and Fahad~Shahbaz Khan.
\newblock Mobillama: Towards accurate and lightweight fully transparent gpt.
\newblock \emph{arXiv preprint arXiv:2402.16840}, 2024.

\bibitem[Tong et~al.(2022)Tong, Song, Wang, and Wang]{tong2022videomae}
Zhan Tong, Yibing Song, Jue Wang, and Limin Wang.
\newblock Videomae: Masked autoencoders are data-efficient learners for self-supervised video pre-training.
\newblock \emph{Advances in neural information processing systems}, 35:\penalty0 10078--10093, 2022.

\bibitem[Toshev and Szegedy(2014)]{toshev2014deeppose}
Alexander Toshev and Christian Szegedy.
\newblock Deeppose: Human pose estimation via deep neural networks.
\newblock In \emph{Proceedings of the IEEE conference on computer vision and pattern recognition}, pages 1653--1660, 2014.

\bibitem[Wang et~al.(2023)Wang, Yang, and Wei]{wang2023learning}
Liang Wang, Nan Yang, and Furu Wei.
\newblock Learning to retrieve in-context examples for large language models.
\newblock \emph{arXiv preprint arXiv:2307.07164}, 2023.

\bibitem[Wang et~al.(2018)Wang, Girshick, Gupta, and He]{wang2018non}
Xiaolong Wang, Ross Girshick, Abhinav Gupta, and Kaiming He.
\newblock Non-local neural networks.
\newblock In \emph{Proceedings of the IEEE conference on computer vision and pattern recognition}, pages 7794--7803, 2018.

\bibitem[Wang et~al.(2022)Wang, Li, Li, He, Huang, Zhao, Zhang, Xu, Liu, Wang, Xing, Chen, Pan, Yu, Wang, Wang, and Qiao]{wang2022internvideo}
Yi~Wang, Kunchang Li, Yizhuo Li, Yinan He, Bingkun Huang, Zhiyu Zhao, Hongjie Zhang, Jilan Xu, Yi~Liu, Zun Wang, Sen Xing, Guo Chen, Junting Pan, Jiashuo Yu, Yali Wang, Limin Wang, and Yu~Qiao.
\newblock Internvideo: General video foundation models via generative and discriminative learning, 2022.

\bibitem[Wu et~al.(2023)Wu, Zhu, Albayrak, Axon, Bertsch, Deng, Ding, Guo, Gururaja, Kuo, et~al.]{wu2023llms}
Tongshuang Wu, Haiyi Zhu, Maya Albayrak, Alexis Axon, Amanda Bertsch, Wenxing Deng, Ziqi Ding, Bill Guo, Sireesh Gururaja, Tzu-Sheng Kuo, et~al.
\newblock Llms as workers in human-computational algorithms? replicating crowdsourcing pipelines with llms.
\newblock \emph{arXiv preprint arXiv:2307.10168}, 2023.

\bibitem[Xu et~al.(2021)Xu, Huang, and Liu]{xu2021sutdtrafficqa}
Li~Xu, He~Huang, and Jun Liu.
\newblock Sutd-trafficqa: A question answering benchmark and an efficient network for video reasoning over traffic events, 2021.

\bibitem[Yang et~al.(2022)Yang, Miech, Sivic, Laptev, and Schmid]{yang2022tubedetr}
Antoine Yang, Antoine Miech, Josef Sivic, Ivan Laptev, and Cordelia Schmid.
\newblock Tubedetr: Spatio-temporal video grounding with transformers.
\newblock In \emph{Proceedings of the IEEE/CVF Conference on Computer Vision and Pattern Recognition}, pages 16442--16453, 2022.

\bibitem[Yang et~al.(2003)Yang, Chaisorn, Zhao, Neo, and Chua]{yang2003videoqa}
Hui Yang, Lekha Chaisorn, Yunlong Zhao, Shi-Yong Neo, and Tat-Seng Chua.
\newblock Videoqa: question answering on news video.
\newblock In \emph{Proceedings of the eleventh ACM international conference on Multimedia}, pages 632--641, 2003.

\bibitem[Yang et~al.(2023)Yang, Chen, Qian, Madaan, Iyengar, Fouhey, and Chai]{yang2023llm}
Jianing Yang, Xuweiyi Chen, Shengyi Qian, Nikhil Madaan, Madhavan Iyengar, David~F Fouhey, and Joyce Chai.
\newblock Llm-grounder: Open-vocabulary 3d visual grounding with large language model as an agent.
\newblock \emph{arXiv preprint arXiv:2309.12311}, 2023.

\bibitem[Yilmaz et~al.(2006)Yilmaz, Javed, and Shah]{yilmaz2006object}
Alper Yilmaz, Omar Javed, and Mubarak Shah.
\newblock Object tracking: A survey.
\newblock \emph{Acm computing surveys (CSUR)}, 38\penalty0 (4):\penalty0 13--es, 2006.

\bibitem[Yu et~al.(2017)Yu, Yin, and Zhu]{yu2017spatio}
Bing Yu, Haoteng Yin, and Zhanxing Zhu.
\newblock Spatio-temporal graph convolutional networks: A deep learning framework for traffic forecasting.
\newblock \emph{arXiv preprint arXiv:1709.04875}, 2017.

\bibitem[Yu et~al.(2024)Yu, Cho, Yadav, and Bansal]{yu2024self}
Shoubin Yu, Jaemin Cho, Prateek Yadav, and Mohit Bansal.
\newblock Self-chained image-language model for video localization and question answering.
\newblock \emph{Advances in Neural Information Processing Systems}, 36, 2024.

\bibitem[Yu et~al.(2019)Yu, Zhang, Er, Li, Xue, Pang, Lin, Tan, Shi, Li, et~al.]{yu2019cosql}
Tao Yu, Rui Zhang, He~Yang Er, Suyi Li, Eric Xue, Bo~Pang, Xi~Victoria Lin, Yi~Chern Tan, Tianze Shi, Zihan Li, et~al.
\newblock Cosql: A conversational text-to-sql challenge towards cross-domain natural language interfaces to databases.
\newblock \emph{arXiv preprint arXiv:1909.05378}, 2019.

\bibitem[Zeng et~al.(2020)Zeng, Xu, Huang, Chen, Tan, and Gan]{zeng2020dense}
Runhao Zeng, Haoming Xu, Wenbing Huang, Peihao Chen, Mingkui Tan, and Chuang Gan.
\newblock Dense regression network for video grounding.
\newblock In \emph{Proceedings of the IEEE/CVF Conference on Computer Vision and Pattern Recognition}, pages 10287--10296, 2020.

\bibitem[Zhang et~al.(2023)Zhang, Sun, Jing, and Zhou]{zhang2023temporal}
Hao Zhang, Aixin Sun, Wei Jing, and Joey~Tianyi Zhou.
\newblock Temporal sentence grounding in videos: A survey and future directions.
\newblock \emph{IEEE Transactions on Pattern Analysis and Machine Intelligence}, 2023.

\bibitem[Zhang et~al.(2020)Zhang, Zhao, Zhao, Wang, Liu, and Gao]{zhang2020does}
Zhu Zhang, Zhou Zhao, Yang Zhao, Qi~Wang, Huasheng Liu, and Lianli Gao.
\newblock Where does it exist: Spatio-temporal video grounding for multi-form sentences.
\newblock In \emph{Proceedings of the IEEE/CVF Conference on Computer Vision and Pattern Recognition}, pages 10668--10677, 2020.

\bibitem[Zheng et~al.(2023)Zheng, Wu, Chen, Yang, Zhu, Shen, Kehtarnavaz, and Shah]{zheng2023deep}
Ce~Zheng, Wenhan Wu, Chen Chen, Taojiannan Yang, Sijie Zhu, Ju~Shen, Nasser Kehtarnavaz, and Mubarak Shah.
\newblock Deep learning-based human pose estimation: A survey.
\newblock \emph{ACM Computing Surveys}, 56\penalty0 (1):\penalty0 1--37, 2023.

\bibitem[Zhou et~al.(2020)Zhou, Neubig, Gu, Diab, Guzman, Zettlemoyer, and Ghazvininejad]{zhou2020detecting}
Chunting Zhou, Graham Neubig, Jiatao Gu, Mona Diab, Paco Guzman, Luke Zettlemoyer, and Marjan Ghazvininejad.
\newblock Detecting hallucinated content in conditional neural sequence generation.
\newblock \emph{arXiv preprint arXiv:2011.02593}, 2020.

\end{thebibliography}

\end{document}